\documentclass[conference]{IEEEtran}
\IEEEoverridecommandlockouts
\usepackage{cite}
\usepackage{multirow}
\usepackage{amsmath,amssymb,amsfonts}
\usepackage{algorithmic}
\usepackage{graphicx}
\usepackage{textcomp}
\usepackage{xcolor}
\def\BibTeX{{\rm B\kern-.05em{\sc i\kern-.025em b}\kern-.08em
    T\kern-.1667em\lower.7ex\hbox{E}\kern-.125emX}}
\begin{document}

\title{Multi-modal Hate Speech Detection using Machine Learning \\
}
\author{\IEEEauthorblockN{Fariha Tahosin Boishakhi}
\IEEEauthorblockA{\textit{Computer Science and Engineering} \\
\textit{BRAC University}\\
Dhaka , Bangladesh \\
fariha.tahosin.boishakhi@g.bracu.ac.bd}
\and
\IEEEauthorblockN{Ponkoj Chandra Shill}
\IEEEauthorblockA{\textit{Computer Science and Engineering} \\
\textit{BRAC University}\\
Dhaka, Bangladesh \\
ponkoj.chandra.shill@g.bracu.ac.bd }
\and
\IEEEauthorblockN{Md. Golam Rabiul Alam}
\IEEEauthorblockA{\textit{Computer Science and Engineering} \\
\textit{BRAC University}\\
Dhaka , Bangladesh \\
rabiul.alam@bracu.ac.bd}}
\maketitle

\begin{abstract}
With the continuous growth of internet users and media content, it is very hard to track down hateful speech in audio and video. Converting video or audio into text does not detect hate speech accurately as human sometimes uses hateful words as humorous or pleasant in sense and also uses different voice tones or show different action in the video. The state-of-the-art hate speech detection models were mostly developed on a single modality. In this research, a combined approach of multi-modal system has been proposed to detect hate speech from video contents by extracting feature images, feature values extracted from the audio, text and used machine learning and Natural language processing. 
\end{abstract}


\section{Introduction}
 In this era of digital communications, hatred data is not only in social media comments and posts or text messages but also in voice messages and video content as well\cite{b1}. Content like this causes cyberbullying, rioting, fraud, loss of respect, and even murder. (Hate Crime Statistics, 2019) shows 15588 law enforcement agencies reported crimes, suspects, offenders, and hate crime zones. These groups reported 7314 hate crimes with 8,559 offenses. The findings include 57.6\%race/ethnicity/ancestry/bias, 20.1 \% religion, and 16.7 \% sexual orientation.\cite{b2} In another research, online hate crimes frequently start online and affect us offline. They were also mistreated online, according to the report's victims. The research shows that hate data is linked to voice tone and facial emotions. Voice tones and facial expressions are crucial aspects to discern hatred. In some cases, only the text data, facial expressions or vocal information as audio data is not enough individually to detect the hateful conversations. Almost all current research on hate speech identification uses text data. This study suggests integrating audio, video, and text elements to detect hate speech. The following research is implemented by taking all the modes available to deliver hate speech into account and designed to detect hate speech more precisely. The final conclusion of hate speech is determined by merging the results of a hard voting ensemble or majority voting where we combine all model result image, audio, and text to determine the final output. 

The major contributions of this research can be summarized as follows - 
\begin{itemize}

\item Data has been prepared with both hateful and non-hateful speech video from different sources such as YouTube, EMBY mostly taken from movies or series. Afterwards, image, audio and text data has been extracted from  the video contents.

\item Feature Extracted from images, audio and text individually. For feature selection, Recursive Feature Selection (RFE), Maximum Relevance - Minimum Redundancy (MRMR) is used to take the most relevant features. 

\item A hate speech detection model has been developed with images, audio and text separately and then combining the results with a hard voting ensemble model to determine the final outcome of hate speech.

\item The performance of seven different classifiers has been studied on the hate speech data set to show the comparative study of different classical machine learning models in hate speech detection.
\end{itemize}
\begin{figure*}[!htbp]
\centering
\includegraphics[width = \textwidth]{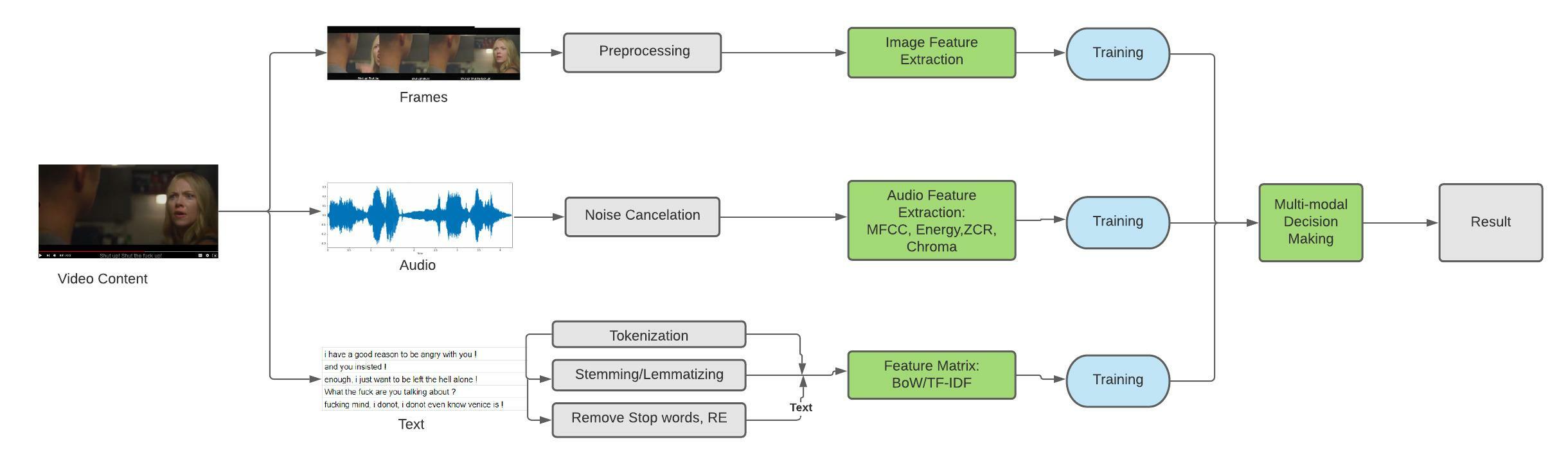} 
\caption{Methodology for Detecting Hate Speech}
\label{Methodology}
\end{figure*}

\section{Literature Review}
Warner and Hirschberg \cite{b5}, for detecting hate speech from websites containing hateful words, epithets, phrases, stereotypical thought, anti-racial content used SVM and with an accuracy of 94\% in classifying anti-semantic speech. The dataset was collected from Yahoo!! and American Jewish Congress.  
\\Kshirsagar et al.(2018) \cite{b8} uses deep learning to categorize online hate speech in general as well as a specific racist and sexist speech using pre-trained word embedding and max/mean from basic fully connected embedding transformations. A new data set that better represents nuanced kinds of hate speech is recommended by the research.
\\
Arora and Singh (2012), mentioned three different approaches for speech recognition, which are acoustic-phonetic approach, pattern recognition, and Artificial intelligence and the paper provided a comprehensive survey of research in speech recognition systems \cite{b10}. After investigating an application of deep neural network architectures to do hate speech detection from tweets, Badjatiya et al.(2017) experimented with multiple classifiers like Logistic regression, SVM, Random forest, and deep neural networks\cite{b13}. LSTM along with Random embedding learned from deep neural network models combined with gradient boosted decision trees led to the best accuracy values. Rybach et al(2009)\cite{b14}  worked with some well-known segmentation methods, focusing on the Log-linear model to determine segmentation from audio streams. GIS(Generalization Iterative Scaling) was used for optimization to detect speech using 3-state HMMs for the signal type speech,  non-speech, pure music, and silence presenting a MAP decoder framework for audio segmentation. Among 15s segments, NIST, ASR-based, and MAP, MAP is the best for audio signals.

All the previous work discussed are on single mode or uses pre-trained model such as BERT. Our Approach shows importance of features from multiple mode and train on simple machine learning model to detect hatred. 

\section{Methodology}
The system model of the proposed multi-modal hate speech detection is presented in Fig.~ \ref{Methodology}. After Collecting the video data we are pre-processing the data by converting the data into image, audio, and text and use image resizing audio noise reduction. Then features from each of the data are extracted separately from each of the data. Both time domain and frequency domain features are extracted from audio data. Each of the extracted features are passed into one single algorithm and fit the values of the features. Each of the features, image, audio, and text will be passed separately to find out the accuracy of the model. The model responds to hate and non-hate for each of the data. After that, the prediction decision-making will be a majority voting ensemble to predict the final output which means two or more modes have to be hate to make the final output to be hate. 
\subsection{Data Description and Pre-processing}For our research, we have collected video data of two categories as Hate and Non-Hate.  Basically, the hatred facial expressions are associated with the negative emotions of anger, fear, aggressiveness, dislikes, disgust found in hateful conversations or can be a sense of injury or violence at the extreme level\cite{b11}.  On the other hand, non-hatred or positive emotions are the feeling of pleasant or desirable behaviors such as joy, amusement, satisfaction, amusement etc.\cite{b12}. we have collected data from different sources such as movies, web series that contain hate speeches along with the images of hatred facial expressions and hatred talks. Non-Hate category, we have considered the videos of positive emotions and extracted data in the same way as hatred data. In consequence, we have prepared a total of 1051 video data for our work.
 Among the 1051 video data, we have used 80\% data as train data and 20\% data as test data. The procedures of extraction of each data are as follows: 
\begin{table}[htbp]
\caption{THE DISTRIBUTION OF HATE AND NON-HATE IN THE data}
\centering
\begin{tabular}{llll} 
\hline
 & Label & percentage in data &   \\ 
\hline
 & Hate & 60\% &   \\
 & Non Hate & 40\% &   \\ 
\hline
 &  &  &  
\end{tabular}
\end{table} 
\begin{itemize}
\item \textbf{ Image:} Each video has been taken at a length of 3 to 5 seconds. OpenCV, a python library, has been used to extract the images, and we have taken converted image data of 30 frames per second. The images have been labelled as Hate and Non-Hate according to their contents. 
\item \textbf{ Audio:} As Audio is unstructured data and, by default, audio signals are non-stationary, meaning their properties differ over time (usually rapidly) it is more efficient to break the record into short segments and calculate one (average) strength value per segment. This is also the central principle behind short-term processing. The video data are converted into audio using the python MoviePy library. and we have reduced the background noise and removed the unnecessary noises.
\begin{figure}[htbp]
\centering
\includegraphics[scale=0.15]{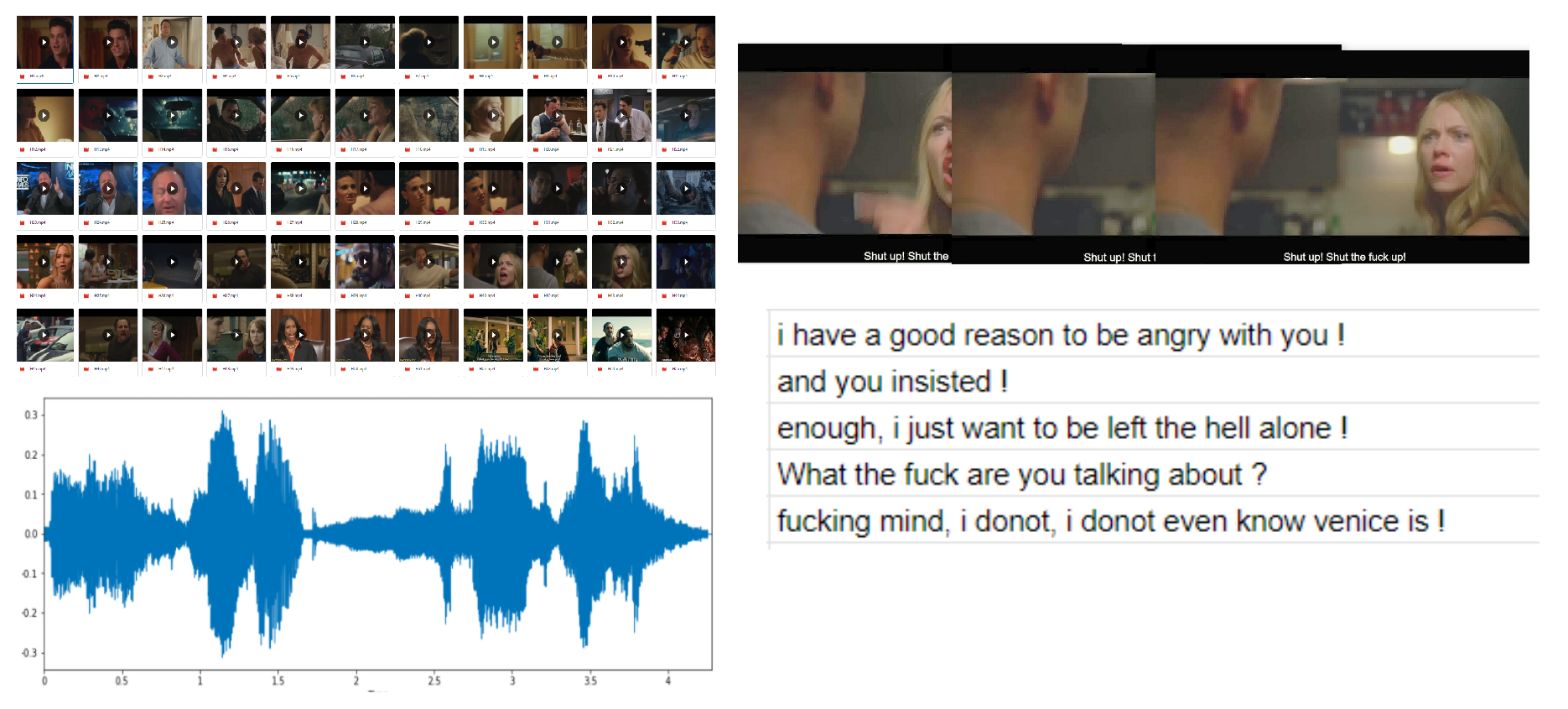} 
\caption{Video,Image, Audio and Text data after pre-processing}
\label{Video,Image, Audio and text data after pre-processing}
\end{figure}
\item \textbf{ Text:} The Audio data were converted into Text using Google Speech Recognition. Each of the data was uploaded to the cloud and converted into text and labelled as Hate and Non-Hate accordingly. Then we have gone through the major text data pre-processing steps such as removing stop words, unnecessary special characters, URLs, double spaces and tokenization has been done for data mining. 
\end{itemize}
\subsection{Feature Extraction and Feature Selection}
\begin{itemize}
\item \textbf{Image:}
For the image data, as the images are categorized into two categories Hate and Non-Hate. We converted the data into pickle files where each of the data is converted into an array.Images of the videos are taken and resized 50 * 50. As the images are nothing but series of array. We are taking all of the images from a video and sorted and label accordingly to extract the image features.  
\end{itemize}
\begin{itemize}
\item \textbf{ Audio:} In the audio data, a sampling rate is determined for the number of samples taken from a continuous signal per second (or per unit) to make a discrete or digital signal. Frequencies are calculated in hertz (Hz) or cycles per second for time-domain signals such as the wave forms for sound (and other audio-visual content types). The Sample rate for the following task is 22050 as we found in Google Speech recognition and they suggested a frequency more than 16 KHz. 
For audio feature extraction two types of features are extracted from the audio signal.

\textbf{Time domain features:} Time domain features are extracted from audio. we took hop length of 256 and frame length of 512 to divide the signal for sampling. From each of the frames we took the following feature and Calculated the mean of the output .  

\textbf{ Energy }can be described as the signal's strength. It is possible to calculate the signal power by its energy. Energy is determined by the sum of signal squares, normalized by frame values the signal's duration.
\begin{equation}
      E(i) = \sum_{n=1}^{W_{L} } |x_{i}({n})|^{2}
\end{equation}
\textbf{ Zero-Crossing Rate } is the rate of signal sign-changes during the frame. In other terms, it is the number of times that the
Signals shift the value, divided by the frame length, from positive to negative and vice versa.\\
\textbf{Frequency Domain Features:}\\
\textbf{Spectral Centroid} is the spectrum's 'gravity' nucleus. It indicates where the” center of mass” for a sound is located and is calculated  as the weighted mean of the frequencies present in the sound.\textbf{ Spectral Rolloff } measures the frequency below which a specified (usually 90\%)  percentage of the total spectral energy.
\textbf{ Mel Frequency Cepstral Coefficient  -} are a limited group of features that describe the overall shape of a spectral envelope. MFCC uses triangular overlapping windows to map the powers of the spectrum onto the mel scale. It takes the discrete cosine transform of the list of mel log powers as if it were a signal. The MFCCs are the resultant spectrum amplitudes.
\textbf{Chroma Vector} is a 12-element representation of the spectral energy. The chroma  vector is computed by grouping the DFT coefficients of a short-term window into 12 bins. Each  bin represents one of the 12 equal tempered pitch classes of the input data. Each bin produces the  mean of log-magnitudes of the respective DFT coefficients.
Entropy of Energy , Spectral Spread, Spectral  Entropy, Spectral Flux are also extracted.

\end{itemize}
\begin{itemize}
\item \textbf{ Text: }First of all, tokenization steps are taken to tokenize the speech into single sentences.In this step we use sentence tokenizer. After tokenizing the whole speech or sentences we eliminate some regular expressions or special characters (like \. \, \_  \@ \# space etc). We just take uppercase and lowercase alphabet (a-z and A-Z). We make all the words into lowercase this time. 
After pre-processing the data we then used the following ways to convert the data to vectors-\\
\textbf{ Counter Vectorizer:} The model of bag-of-words (BOW) is a representation that transforms arbitrary text into vectors of fixed length by counting the number each word appears. Here, based on the label of the data and each word of a single data or sentence are given values.\\
\textbf{ Term Frequency-Inverse Document Frequency (TF-IDF):} TF-IDF is a statistical test that assesses in a series of documents how important a term is to a text. 
The formula that is used to compute the tf-idf is : 
\begin{equation}
    TF-IDF = TF * IDF
\end{equation}    
    where,    
    TF = (Number of time the word occurs in the text) / (Total number of words in text)
IDF = log(Total number of documents / Number of documents with word t in it)
\end{itemize}
After feature extraction, the most relevant characteristics were chosen using Recursive Feature Selection (RFE) and Maximum Relevance - Minimum Redundancy (mRMR). Recursive Feature Selection is a feature selection approach that wraps around a filter-based feature selection process. RFE works by searching for a subset of features in the training data, starting with all of the features and successfully deleting them until the target number remains. Maximum Relevance - Minimum Redundancy is a "minimum optimum" feature selection technique that aims to get the greatest possible prediction performance from a group of features given a certain number of features. Entropy of Energy , Spectral Spread, Spectral  Entropy, Spectral Flux are eliminated by the feature selection.

\section{EVALUATION AND RESULTS }
Following machine learning models have been used for evaluation of the performance:
\textbf{Support Vector Machine }finds the decision boundary that optimizes the distance from the nearest data points.
\textbf{Random Forest } algorithm produces decision trees on data samples and then gets the prediction from each of them and selects the best solution by voting. It is an ensemble technique that is better than a single decision tree because by combining the result, it decreases the over-fitting.
\textbf{Logistic Regression:} A statistical model used to evaluate if an independent variable has an effect on a binary dependent variable is logistic regression. This implies that given an input, there are only two possible results.
\textbf{AdaBoost Classifier} is a meta-estimator that starts with one classifier and adds more to the same data set, but modifies the weights of wrong classifications to focus on more complex situations later. Adaboost\cite{b20} arbitrarily picks a training subset. A training set is chosen based on a forecast for the Adaboost machine learning model's last training. It lends additional weight to incorrectly classified observations so they are categorised in the next rotation.
 \textbf{ K-nearest neighbors} is a Lazy algorithm implies that model generation does not require any training data points. The core decision factor is the number of neighbors. K is normally an odd number, because the number of classes is 2.
\textbf{ Naive Bayes Classifier} are a set of Bayes Theorem-based classification algorithms.It is a grading methodology based on the Bayes theorem, which implies that predictors are independent.The Gaussian model assumes that a normal distribution is followed by features. This means that the model assumes that the values are sampled from the Gaussian distribution if predictors take continuous values instead of discrete ones.
\textbf{ Decision Tree} starts with a single node that splits into possible outcomes. Any result leads to additional nodes that connect with other alternatives. It resembles a tree. An automated statistical model for machine learning, data processing, and statistics may be built using a decision tree.
\textbf{ Multi-modal approach }After all the individual classification done though the classical machine learning algorithm, a ensemble learning hard voting method or majority voting is used to get the final output combining the three individual detection. If the detection shows two or more than two individual outputs are positive to hate then the video is detected as hate otherwise it is detected as non hate speech.
\\

\begin{table}[]
\centering
\caption{Result Analysis For Image, Audio, Text, and Multi-modal}
\label{tab:Result analysis}
\begin{tabular}{|l|l|l|l|l|}
\hline 
Algorithm & Data & Precision & Recall & F1Score \\ \hline

\multirow{4}{*}{SVM} & Image & 0.8959 & 0.8972 & 0.8965 \\ \cline{2-5} 
                     & Audio & 0.8047 & 0.81   & 0.84   \\ \cline{2-5} 
                     & Text  & 0.775  & 0.813  & 0.793  \\ \cline{2-5} 
                     & Multi-modal & 0.74 & 0.813 & 0.875 \\ \hline
                  
\multirow{4}{*}{Random Forest} & Image       & 0.80  & 0.80  & 0.80  \\ \cline{2-5} 
                               & Audio       & 0.874 & 0.875 & 0.874 \\ \cline{2-5} 
                               & Text        & 0.841 & 0.70  & 0.764 \\ \cline{2-5} 
                               & Multi-modal & 0.65  & 0.81  & 0.718 \\ \hline
                  
\multirow{4}{*}{Logistic Regression} & Image       & 0.897 & 0.8976 & 0.8973 \\ \cline{2-5} 
                                     & Audio       & 0.874 & 0.80   & 0.887  \\ \cline{2-5} 
                                     & Text        & 0.740 & 0.80   & 0.851  \\ \cline{2-5} 
                                     & Multi-modal & 0.708 & 0.85   & 0.772  \\ \hline
                                     
\multirow{4}{*}{Adaboost} & Image       & 0.841 & 0.82  & 0.787 \\ \cline{2-5} 
                          & Audio       & 0.86  & 0.848 & 0.874 \\ \cline{2-5} 
                          & Text        & 0.81  & 0.571 & 0.727 \\ \cline{2-5} 
                          & Multi-modal & 0.87  & 0.761 & 0.780 \\ \hline
                          
\multirow{4}{*}{k-NN} & Image       & 0.828 & 0.797 & 0.777 \\ \cline{2-5} 
                      & Audio       & 0.787 & 0.748 & 0.760 \\ \cline{2-5} 
                      & Text        & 0.818 & 0.857 & 0.837 \\ \cline{2-5} 
                      & Multi-modal & 0.8   & 0.761 & 0.780 \\ \hline
                      
\multirow{4}{*}{Naive Bayes} & Image       & 0.736 & 0.858  & 0.792 \\ \cline{2-5} 
                             & Audio       & 0.804 & 0.827  & 0.85  \\ \cline{2-5} 
                             & Text        & 0.707 & 0.7333 & 0.709 \\ \cline{2-5} 
                             & Multi-modal & 0.756 & 0.846  & 0.784 \\ \hline
                             
\multirow{4}{*}{Decision Tree} & Image       & 0.79 & 0.80 & 0.85 \\ \cline{2-5} 
                               & Audio       & 0.726 & 0.704 & 0.715 \\ \cline{2-5} 
                               & Text        & 0.775 & 0.713 & 0.793 \\ \cline{2-5} 
                               & Multi-modal & 0.74  & 0.813 & 0.775 \\ \hline

\end{tabular}
\end{table}

From table \ref{tab:Result analysis} we can see that the multi modal approach performed well on AdaBoost and Naive Bayes where both achieved accordingly 87\% and 75\% accuracy than image, text and audio individually. SVM also performed 74\% accuracy and 81.3\% precision which is less than each of the mode separately.    
\section{CONCLUSION}
The spread of hate must be stopped. This research detected hate and offensive content by analyzing audio, text, and video. The studied machine learning models achieved the highest 87\% accuracy using the carefully extracted discriminative features.

\end{document}